\pdfoutput=1
\documentclass{sig-alternate}
\usepackage{multirow}

\begin{document}
%

\conferenceinfo{GECCO'13,} {July 6--10, 2013, Amsterdam, The Netherlands.}
\CopyrightYear{2013}
\crdata{978-1-4503-1963-8/13/07}
\clubpenalty=10000
\widowpenalty = 10000

\title{Self Organizing Classifiers and Niched Fitness}

%
%
%
%
%

\author{
%
%
\alignauthor 
Danilo V. Vargas\\  
       \affaddr{Graduate School of Information Science and Electrical Engineering,\\Kyushu University}\\
       \affaddr{Fukuoka, Japan}\\
       \email{vargas@cig.ees.kyushu-u.ac.jp}
\alignauthor 
Hirotaka Takano\\ 
       \affaddr{Faculty of Information Science and Electrical Engineering, Kyushu University}\\
       \affaddr{Fukuoka, Japan}\\
       \email{takano@cig.ees.kyushu-u.ac.jp}
\alignauthor 
Junichi Murata\\ 
       \affaddr{Faculty of Information Science and Electrical Engineering, Kyushu University}\\
       \affaddr{Fukuoka, Japan}\\
       \email{murata@cig.ees.kyushu-u.ac.jp}
}

\numberofauthors{3} 
%
\date{30 July 1999}

\maketitle
\begin{abstract}
Learning classifier systems are adaptive learning systems which have been widely applied in a multitude of application domains. 	
However, there are still some generalization problems unsolved.
The hurdle is that fitness and niching pressures are difficult to balance.
Here, a new algorithm called Self Organizing Classifiers is proposed which faces this problem from a different perspective.
Instead of balancing the pressures, both pressures are separated and no balance is necessary. 
In fact, the proposed algorithm possesses a dynamical population structure that self-organizes itself to better project the input space into a map.	
The niched fitness concept is defined along with its dynamical population structure, both are indispensable for the understanding of the proposed method.
Promising results are shown on two continuous multi-step problems.
One of which is yet more challenging than previous problems of this class in the literature.
\end{abstract}

\category{I.2.11}{Artificial Intelligence}{Distributed Artificial Intelligence}[intelligent agents, languages and structures, multiagent systems]
\category{I.2.6}{Artificial Intelligence}{Learning}

\terms{Algorithms, Performance}

\keywords{Self Organization, Self Organizing Systems, Self Organizing Map, Learning Classifier Systems, Reinforcement Learning, Structured Evolutionary Algorithms}

\section{Introduction}


The importance of classifiers to maximize their reward (decrease the error in supervised learning problems or increase reward in reinforcement learning problems) with at the same time specializing to their best suited position (niching) were demonstrated by many articles about learning classifier systems (LCS) \cite{urbanowicz2009learning, lanzi2000roadmap}.
Throughout the LCS history, there has been an initial focus on strength based systems (e.g., ZCS \cite{wilson1994zcs}) which highlight the focus on reward. Later, however, it was realized that much of the niching capability may be lost in strength based systems, because over-general classifiers dominate and high payoff niches become crowded.
Accuracy based systems, on the other hand, (e.g., XCS \cite{wilson1995classifier, butz2001algorithmic}, XCSF \cite{wilson2002classifiers}) try a middle ground between niching and reward.
By rewarding the accuracy of the reward prediction, accuracy based systems have justifiably better niching than strength based systems (the accuracy-based fitness is more uniform over different niches). 
For this reason, recent articles have focused more on accuracy-based systems.

However, it is known for some time that XCS systems (and in general, accuracy-based systems) may have problems with generalization \cite{lanzi1997study, butz2004toward}.
Solutions have been proposed for some environments \cite{lanzi2007generalization}, but the hindrance is that different pressures are acting together and balancing them properly for every environment is difficult \cite{horn1994implicit}.
In other words, the balance between specialized (niched) and generic classifiers is still unsolved.

This article propose to face the dilemma from a different perspective. 
What if we separate completely the niching problem from the fitness problem?
Suppose we could divide the input space into small niches \footnote{
Concerning the proposed method, the term niche follows the Hutchinsonian niche definition \cite{Hutchinson1957}.
Hutchinsonian niche is an $n$-dimensional hyper-volume composed of environmental features.
} and evolve the solutions for each niche.
By doing this, the fitness for each niche can be strength based because the niches are small and independently defined (not defined by the chromosome and thus not affected by evolution).
Moreover, there is not any balance necessary, since niching and fitness pressures were separated from each other.

But then, two questions need to be solved:
\begin{enumerate}
	\item How can we divide the input space automatically and compactly?
	\item How can we compare individuals? Notice that the fitness defined above is relative to the niche!
\end{enumerate}

The answer to the first question lies in using Self Organizing Map (SOM).
Self Organizing Map is an unsupervised artificial neural network \cite{kohonen1990self}.
It is an algorithm capable of projecting high dimensional spaces into a map. 
In other words, by using SOM, a multidimensional input space can be divided automatically into the cells of a compact map. 
In fact, the projection done preserves the topological properties and retains many other relevant information (e.g., data frequency) of the input space.

The second question has a simple answer: they can not be compared.
With relative fitness, there is not any easy way of comparing individuals from different niches.
Therefore, regarding the fitness selection, the evolutionary algorithm (EA) should be constrained to each niche.
Genetic operators, however, are not constrained and may benefit from any population structure present. 

Actually, the proposed approach is evaluated at two continuous multi-step problems with promising results.
The experiments conducted on two continuous labyrinths demonstrate the capabilities of the proposed approach.
One of which is more challenging than the continuous multi-step problems solved by state of the art LCS. 

\section{Learning Classifier Systems in Multi-step and/or Continuous\\ Problems}

Learning classifier systems are evolutionary based systems capable of solving problems by the mutual competition and/or cooperation of their solutions.
There is a wide and diverse literature. Here we will confine to a brief review of LCS applied to multi-step and/or continuous problems.
For a detailed review of the literature, please refer to \cite{urbanowicz2009learning, lanzi2000roadmap}.

LCS with continuous actions were applied to many problems.
To begin with, XCSF has been applied to function approximation \cite{wilson2002classifiers, butz2008function,tran2007xcsf}.
Other works in function approximation include the LCS with fuzzy logic  \cite{valenzuela1991fuzzy, bull2002accuracy, casillas2007fuzzy}, neural-based LCS algorithms \cite{bull2002using, bull2002accuracy} and genetic programming-based \cite{iqbal2012xcsr}.
The success of LCS also span the control of robotic arms  \cite{stalph2012learning, butz2008context} and navigation problems \cite{bonarini2000fuzzy,howard2009towards}.

However, applications to multi-step problems with continuous actions restrict to the mobile robot in a corridor \cite{bonarini2000fuzzy} and the empty room with noise \cite{howard2009towards}.
Complex multi-step problems were solved only for \textbf{discrete outputs} \cite{lanzi2005xcs}.

\section{Structured Evolutionary Algorithms}

Structured evolutionary algorithms does not possess a panmictic population.
Instead they organize the individuals into a structured population \cite{tomassini2005spatially, alba2002parallelism}.

Two types of structured EAs will be given as examples which are somewhat related to the structure of the proposed method.

\begin{figure}
\centering
\includegraphics[height=1in]{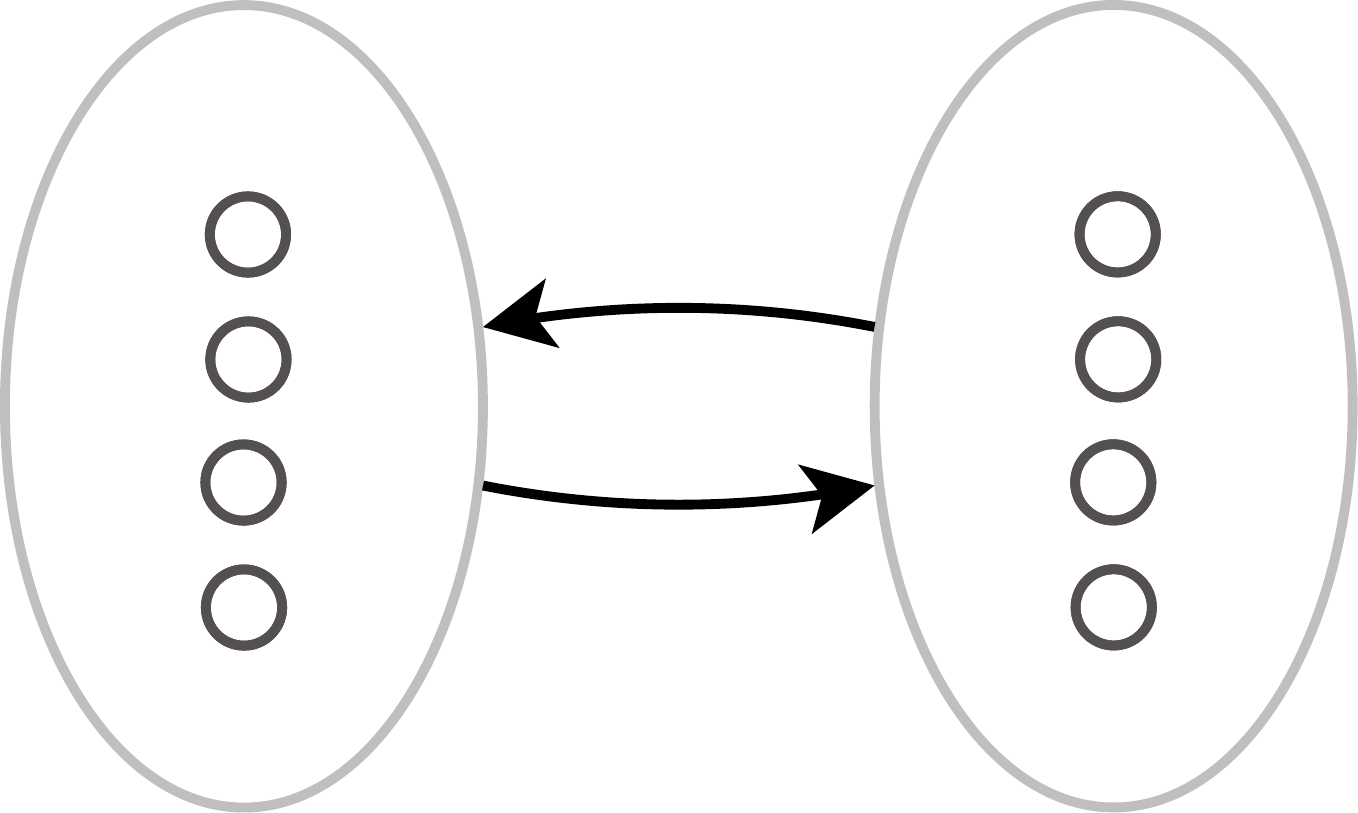}
\caption{Island model structure. Arrows indicate the infrequent immigration procedure, the circles are the individuals and the oval shapes are the subpopulations.}
\label{island}
\end{figure}

The first type is island models (also called distributed genetic algorithms) \cite{belding1995distributed}.
Figure~\ref{island} shows the structure. 
Basically, the population is divided into a number of subpopulations (``islands") with few genetic information exchanged between them.

\begin{figure}
\centering
\includegraphics[height=1in]{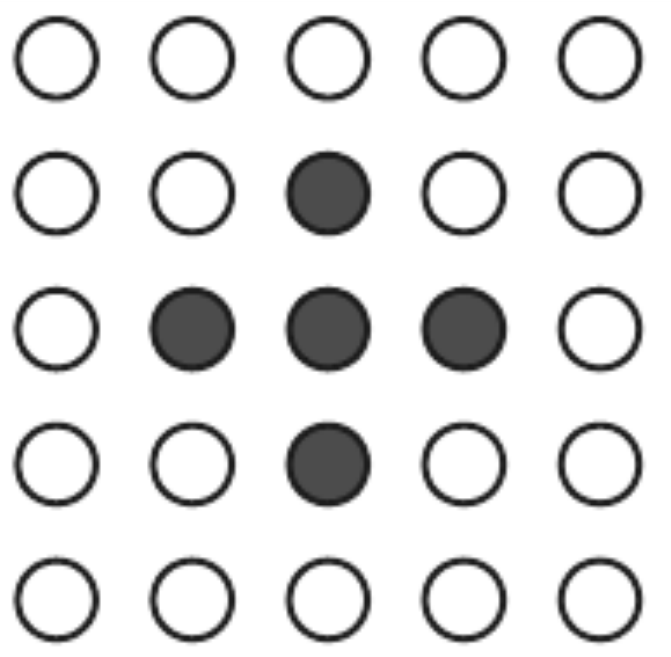}
\caption{Cellular algorithm structure. Shaded area indicates an example of neighborhood for the central individual.}
\label{cellular}
\end{figure}

The second type, cellular algorithms are structured evolutionary algorithms where individuals are usually positioned in a vertex of a lattice graph (Figure~\ref{cellular} shows a common cellular structure). 
They interact solely with adjacent individuals defined by the a neighborhood function \cite{manderick1989, alba2000cellular}.

\section{New Concepts}

First, in order to present the proposed method, it is necessary to explain some new concepts and their related literature.

\subsection{Niched Fitness}

A fitness which is relative to a given place or circumstance is called \textbf{niched fitness}.
Thus, it does not possess a meaning when compared with another fitness out of this place or circumstance.

This is a new concept. 
At first glance, it may appear unwise to use niched fitness, because the comparison between individuals from different niches becomes impossible.
But there is the benefit of avoiding competition between tasks different in nature.
Therefore, every niche existence is protected from other more rewarding and/or more voluminous and/or more frequently accessed niches.

Moreover, niched fitness can be thought as a multiobjectivization procedure \cite{mouret2011novelty}, because each niched fitness becomes a separate fitness.
In fact, niching and multi-objective search are related subjects.
The Pareto front created allows the existence of multiple non-dominated individuals (different niches).
We consider this a relevant aspect which deserves to be mentioned.
The remaining of this article, however, will ignore this aspect and treat the search as a single objective one. 

\subsection{SOM Population}

\begin{figure}
\centering
\includegraphics[height=1.5in]{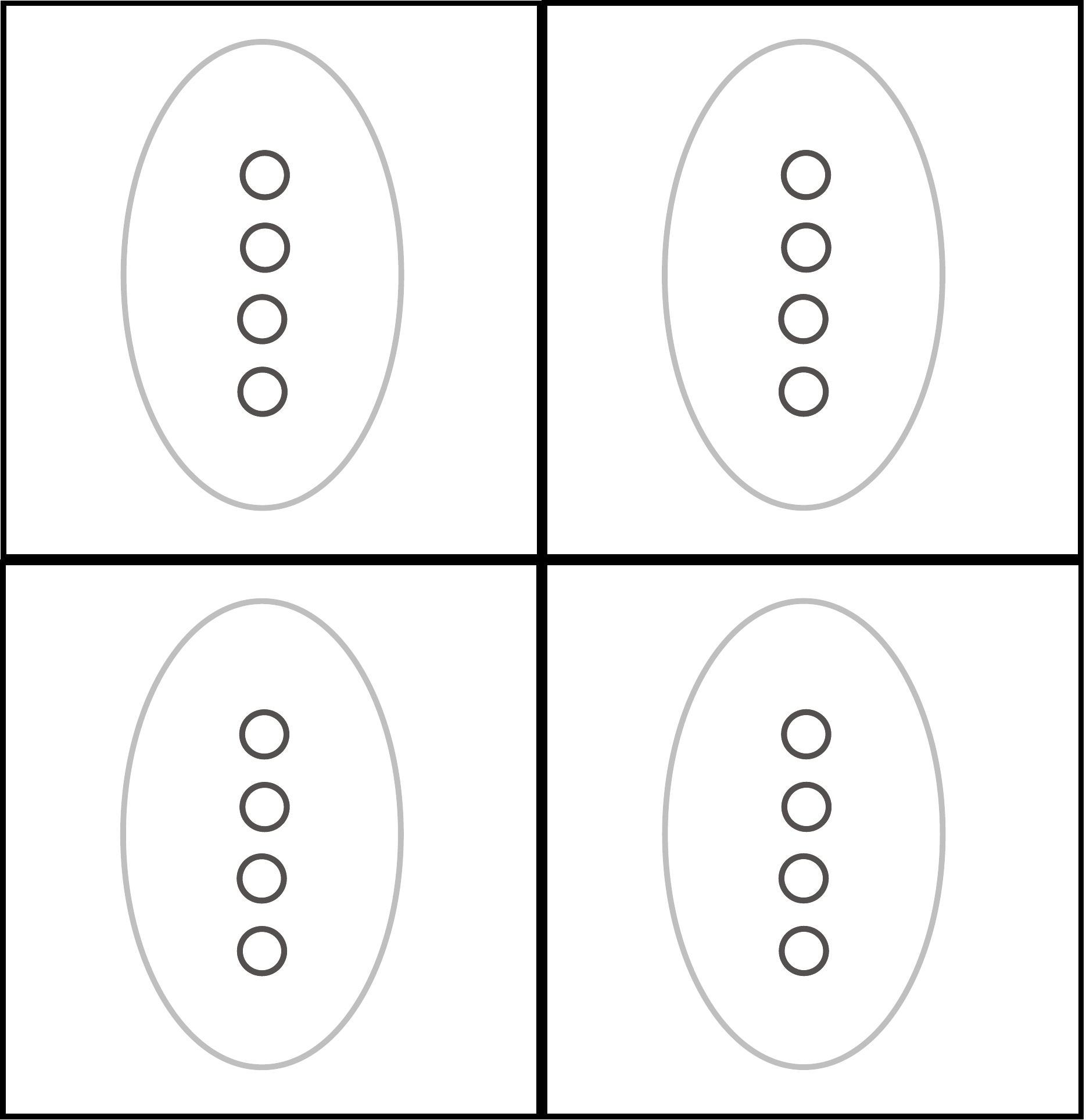}
\caption{SOM population structure. A self organizing map grid with populations inside each cell.}
\label{som_pop}
\end{figure}

This Section aims to define the SOM population concept.
SOM population is a 2D grid with each cell in the grid having a subpopulation.
Moreover, the 2D grid is a projection map which is not static, i.e., it is part of a SOM which is always self-organizing itself.
Figure~\ref{som_pop} illustrates the SOM population (SOM's weight vectors are excluded for simplification).
Generally speaking, this structure can be seen as a mixture of both island models and cellular algorithms, with the additional feature that the structure also changes its properties in relation to the input.

The SOM population's behavior is described below.
When an input is given to a SOM population, a competition takes place between the cells of the map.
The cell which is closer wins the competition to the input and the SOM cells in the neighborhood are updated (the update make the SOM cell's weight vectors get closer to the input).
Then, an individual inside the winner cell acts on the environment and is consequently evaluated by a subsequent reward.
The individual chosen to act inside the winner cell is based on a selection procedure which is algorithm specific.
(here we use a random selection procedure, see Section~\ref{soc}).

\section{Self Organizing Classifiers}
\label{soc}

\begin{figure*}
\centering
\includegraphics[height=3in]{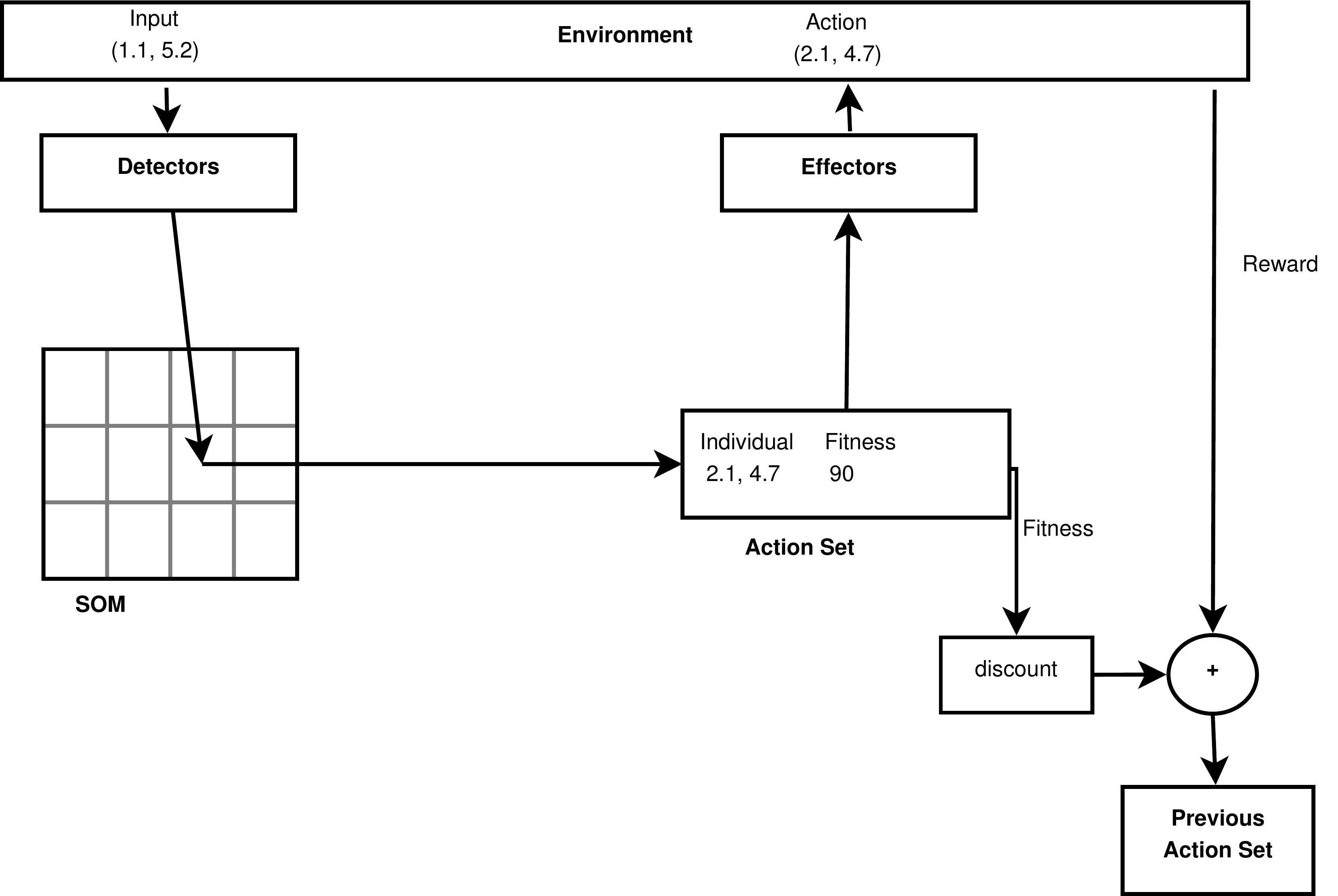}
\caption{Self Organizing Classifier Schematic}
\label{schematic}
\end{figure*}

Figure~\ref{schematic} shows a schematic of the overall procedure of \textbf{Self Organizing Classifiers} (SOC).
It follows the classic schematic style used in ZCS, XCS and many other papers of the LCS literature \cite{wilson1994zcs}.
Simple classifiers are used. 
They are made up of action parameters represented by an array of real numbers. 

SOC uses a Q-learning based reinforcement scheme with niched fitness.
The fitness update of each individual is done using the Widrow-Hoff rule \cite{Widrow1960Adaptive}:
\begin{equation}
F = F + \eta(\hat{F} - F),
\end{equation}
where $\eta$ is the learning rate, $F$ is the current fitness and $\hat{F}$ is a new fitness estimate.
The fitness estimate of cell $cell$ and classifier $c$ which were activated at time $t-1$ is given by the following equation:
\begin{equation}
	\hat{F}(c,cell)_{t-1} = R_{t-1} + \gamma \underset{c' \in cell'}{\operatorname{max}} \{F(c',cell')\},
\end{equation}
where $R$ is the reward received, $\gamma$ is the discount-factor and
$\underset{c \in cell}{\operatorname{max}} \{F(t)\}$ is the maximum fitness of classifier $c'$ inside the activated cell $cell'$ at the current cycle $t$.

Similar to learning classifier systems, to decrease computation resources, the structure of the SOM population is implemented as a single array of classifiers with a given numerosity indexed by the SOM population structure.
In this manner, the numerosity is defined by the number of indexes a given individual possess.

This article uses a particular SOM population where the subpopulation inside each cell is divided into two groups: one of best individuals and the other of novel individuals. 
Best and novel individuals have a fixed size of $\beta$ and $\nu$ respectively.
Considering EA's cycle is an algorithm cycle when the EA is called, the following rules take place:
\begin{itemize}
\item Best individuals are the best fitted individuals inside the subpopulation in the last EA's cycle.
\item Novel individuals are renewed every EA's cycle (the detailed process is described in the next Section). 
\end{itemize}

The SOM population begins without any classifiers. 
Classifiers are created when the respective cell wins the competition inside the SOM.
In one hand, novel individuals are created as random classifiers.
On the other hand, best individuals, when possible, are set equal to another cell's best individuals from the neighborhood\footnote{Neighborhood is defined as the cells within a Chebishev distance of less or equal to four} which maximize $\frac{experience}{chebishevDistance^2}$.
If not possible, best individuals are initialized in the same way as the novel individuals.

The system has, as usual in reinforcement learning, cycles of exploration and exploitation.
Within the SOM's winning cell in a giving exploration or exploitation cycle a random individual from respectively the novel or best individuals are chosen to act.
Moreover, cycles of exploration and exploitation are always alternated (a cycle of exploration is followed by an exploitation cycle and so on).

\subsection{Evolution}

For every cycle that a cell's individual acts, this cell has its experience counter increased.
The evolutionary algorithm is called locally on each cell when the cell's experience is greater than $\iota S$. 
Where $S$ is the number of subpopulation individuals (novel plus best individuals) present on each cell. 
The parameter $\iota$ defines an experience per individual, above which they should have an accurate fitness evaluation.

By applying the evolutionary algorithm locally, it respects the niched fitness concept.
Its procedure consists of sorting the individuals of the given cell according to their fitness.
The current best $\beta$ individuals substitute the previous best individuals and the remaining individuals are discarded (the index is removed and the individual numerosity decrease, if the numerosity reaches $0$ it is deleted).
Novel $\nu$ individuals are created using either:
\begin{enumerate}
	\item Indexing - A copy from (index to) a randomly selected individual of the entire population;
	\item Reproduction - Created by a genetic operator.
\end{enumerate}		
The two procedures above have equal probabilities.

Motivated by some comparison articles and robustness tests, the differential evolution is chosen as the genetic operator \cite{storn1997differential, vesterstrom2004comparative, iorio2005solving}.
It compares well to even complex optimization algorithms (e.g., Estimation of Distribution Algorithms) \cite{garcia2009study}.
The differential evolution's mutant vector is created by randomly choosing three vectors from the SOM's entire population of individuals (individuals with numerosity bigger than one are counted as one).

\section{Experiments}

\subsection{Environments}

\begin{figure}
\centering
\includegraphics[height=1.5in]{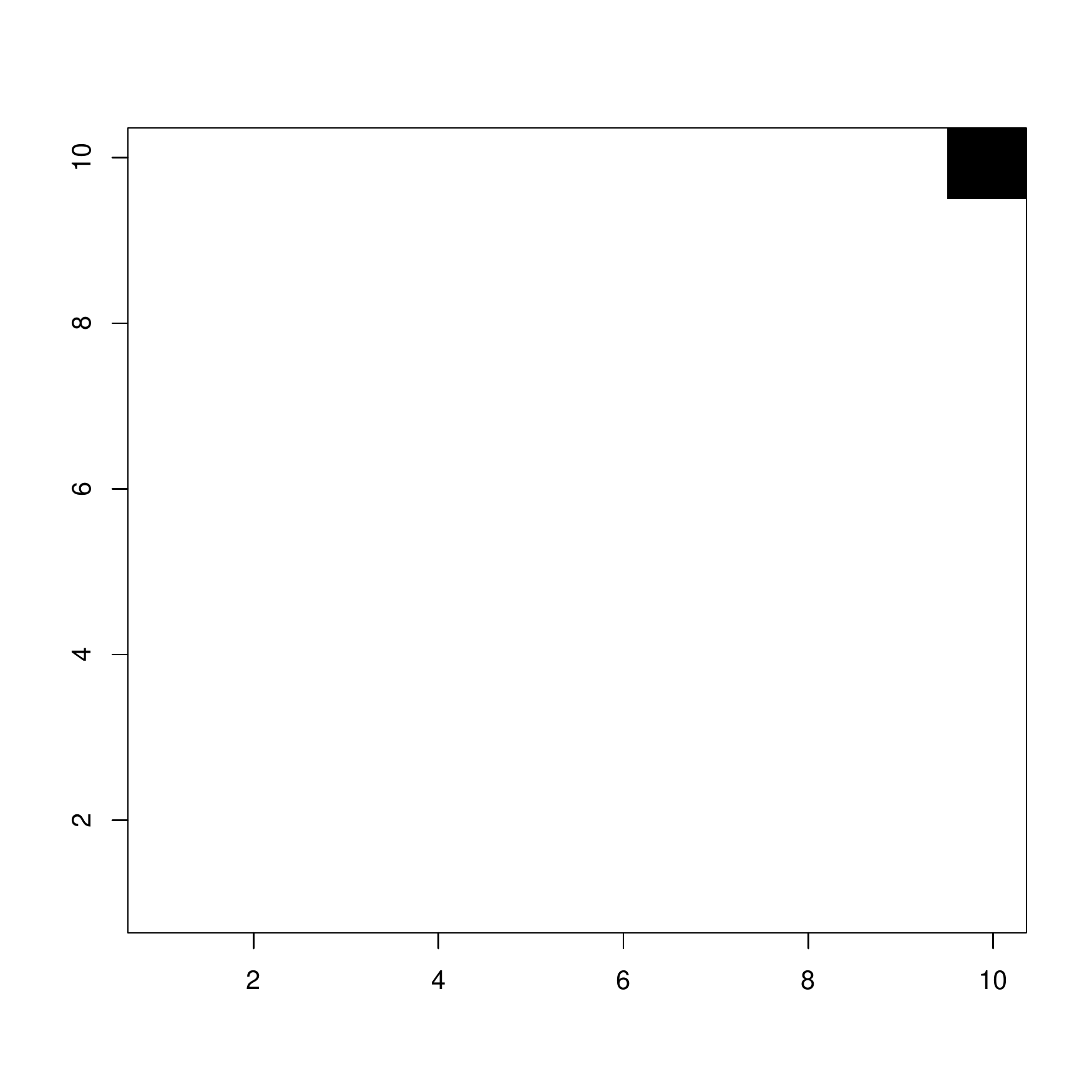}
\caption{Empty room environment. The black position represents the goal.}
\label{p1}
\end{figure}

\begin{figure}
\centering
\includegraphics[height=1.5in]{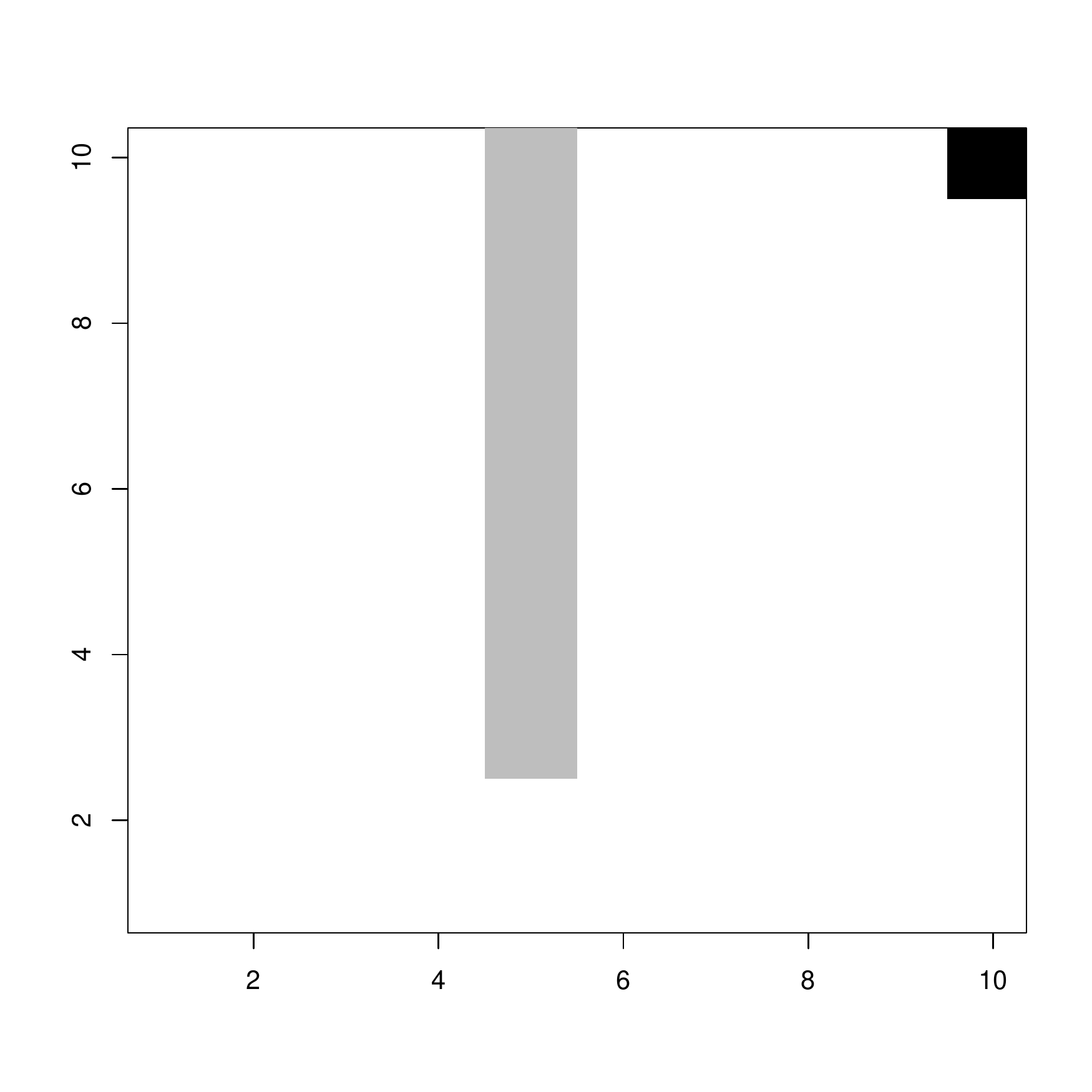}
\caption{One-Wall Maze environment. Grey positions are obstacles and the black position is the goal.}
\label{p2}
\end{figure}

The experiments were conducted on both Empty Room and One-Wall Maze environments.
They are respectively depicted in Figures~\ref{p1} and~\ref{p2}.
All environments require continuous actions. 
The variable observed by the agent is the agent's position which is also continuous. 

At every trial, the agent starts at a random position on the environment. 
Naturally, starting inside a wall is not possible.
Reaching the goal would give the agent a reward of $1000$, hitting an obstacle would return $-20$ and any other action would return $-10$.
Additionally, agents can not move more than $1.0$ in any direction.
The collision system is simply implemented (which makes it harder than a real system). 
If an agent tries to move inside a wall, the system detects the infraction, sets the agent in the previous position and returns the reward.
In other words, an agent constantly hitting the wall will not move at all.
However, an agent that hits the limits of a maze will have its final position limited by the environment.
Therefore, it is possible to move sideways when hitting the limit of the environment.

\subsection{Settings and Design of Experiments}

The parameters of the algorithm are fixed and described in Table~\ref{para}.
Here, $it$ means the SOM iteration number, $chebyshevDistance()$ is the Chebyshev distance between the current cell and the cell which won the SOM's competition and $random(a,b)$ is a function which returns a uniform random value between $a$ and $b$.
The cells of the SOM are only updated if the neighborhood function multiplied by the learning restraint surpasses the cell update threshold.

Following the design of \cite{lanzi2005xcs} the performance is computed as the average steps to reach the goal during the last $100$ trials. 
The trials can not last more than $500$ steps. 
Any trial which last more than $500$ is terminated and a new trial is started with the agent, as usual, in a random position.
All statistics, when not stated otherwise, are averaged over $20$ experiments.

\begin{table*}
\centering
\caption{Parameters}
\begin{tabular}{ |c|l|l| }
	\hline
	\multirow{2}{*}{Differential Evolution} & CR & $0.2$\\
	& F & $random(0,1)$ \\ \hline
	\multirow{5}{*}{Self Organizing Map} & Matrix Size & $10\times10$ \\
	 & Weight's initial value & $random(0,1)$ \\ 
	 & Learning restraint & $0.1(0.999999)^{it}$ \\
	 & Neighborhood function & $\exp(-chebyshevDistance()^2)$ \\
	 & Cell update threshold & $0.005$ \\
	\hline
	\multirow{5}{*}{Self Organizing Classifiers} & $\eta$ & $0.2$ \\ 
	 & $\beta$ & $5$ \\
	 & $\nu$ & $10$ \\
	 & $\iota$ & $20$ \\
	 & $\gamma$ & $0.9$ \\
	 & $Initial Fitness$ & $0$ \\
	\hline
\end{tabular}
\label{para}
\end{table*}

\subsection{Empty Room}

Empty Room is a well known problem, in continuous \cite{howard2009towards} as well as discrete action spaces \cite{lanzi2005xcs}.
The evaluation focus here is on the overall functionality of the system.
Although not complicated in nature, this problem may expose strong instabilities and other undesired phenomena.

\begin{figure}
\centering
\includegraphics[height=2.5in]{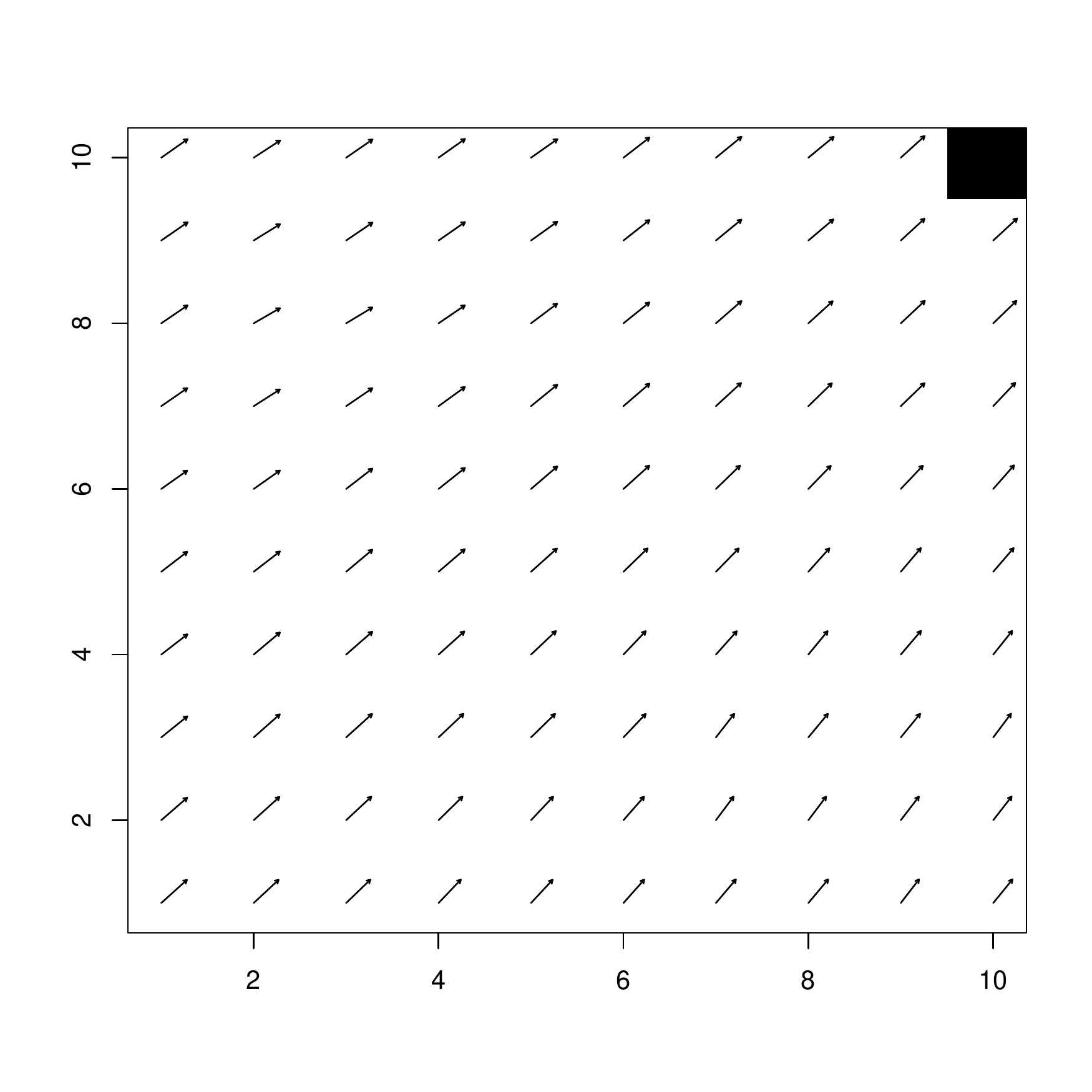}
\caption{Behavior evolved in the Empty Room environment.}
\label{be1}
\end{figure}

Figure~\ref{be1} shows the average behavior obtained.
To construct this figure, the action of the agent was sampled $100$ times and averaged in a given environment area of size $1$x$1$.
The resulting matrix have a dimension of $10$x$10$ averaged actions.
Subsequently, the matrices of $20$ experiments in the same environment were averaged.

\begin{figure}
\centering
\includegraphics[height=2.5in]{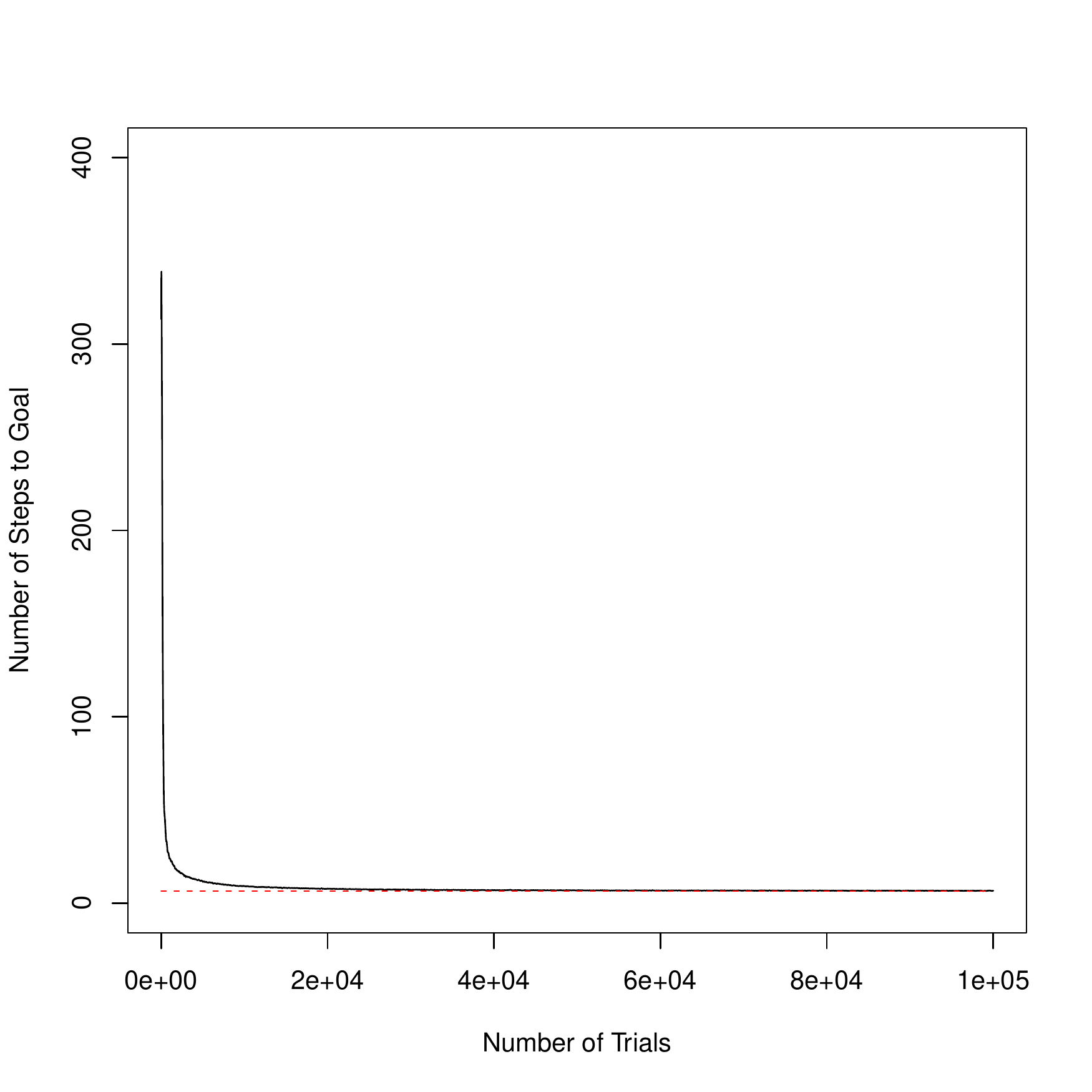}
\caption{Performance over the Empty Room environment (continuous line). For guiding purposes a dashed line is drawn at 6.5 of distance to the x-coordinate.}
\label{pe1}
\end{figure}

\begin{figure}
\centering
\includegraphics[height=2.5in]{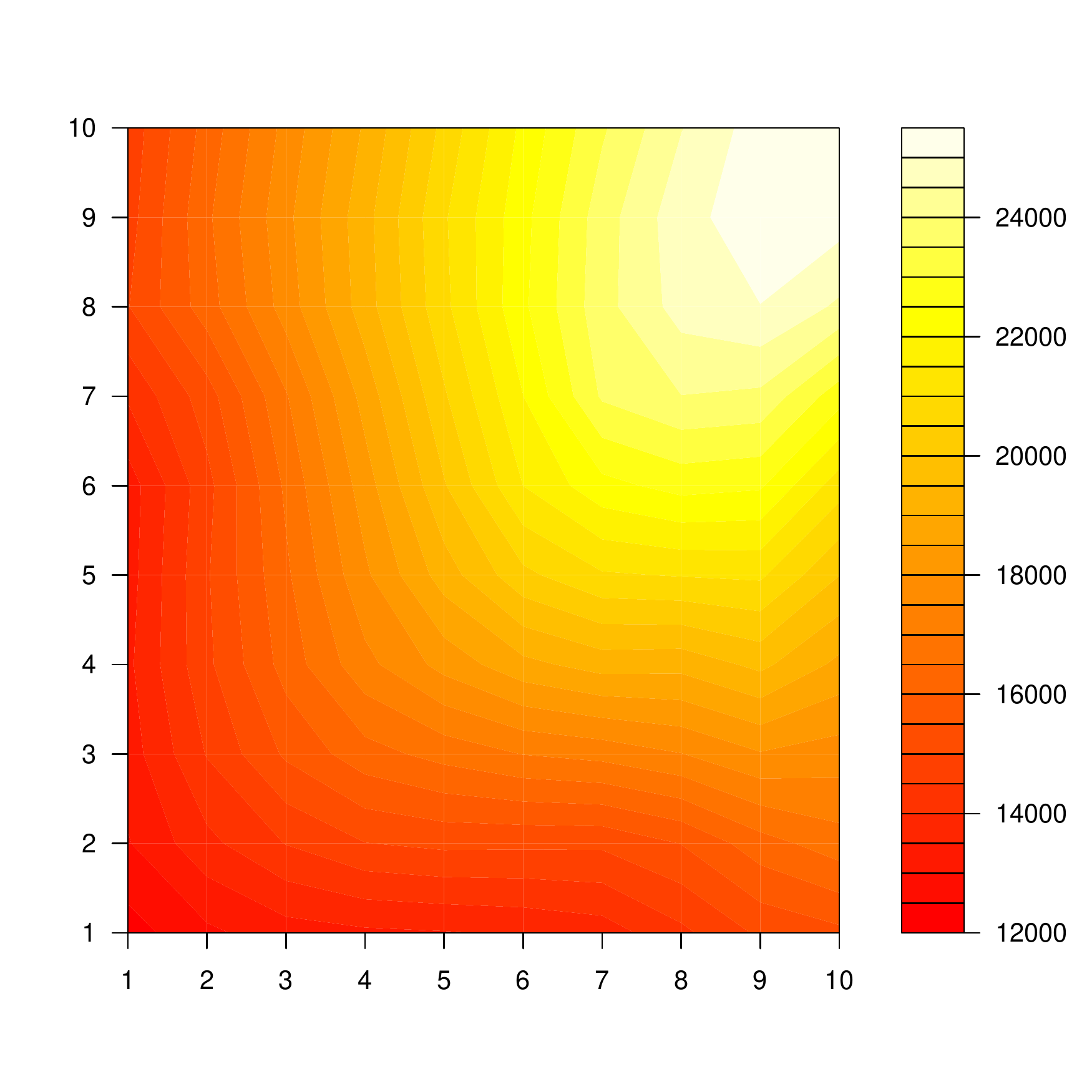}
\caption{Average distribution of fitness in the Empty Room environment.}
\label{fit1}
\end{figure}

The SOC's developed behavior is optimal, in spite of the fact that inside the SOM structure each cell must arrive at the best solution (fitness is not global). 
The justification lies in the fact that the ability of the SOC algorithm to share and improve solutions between cells make the development very fast and stable.
Figure~\ref{pe1} shows the performance.
The fitness distribution is calculated in the same way as the matrix of actions, but instead of action vectors the maximum fitness of the cell is measured.
The result is shown on Figure~\ref{fit1}.

\subsection{One-Wall Maze}

This second problem is different from the previous one. 
It makes necessary the use of different actions for different places of the environment. 
Note that the classifiers used are very simple.
Therefore, this experiment evaluates specifically the dynamical population structure's niching capabilities.

\begin{figure}
\centering
\includegraphics[height=2.5in]{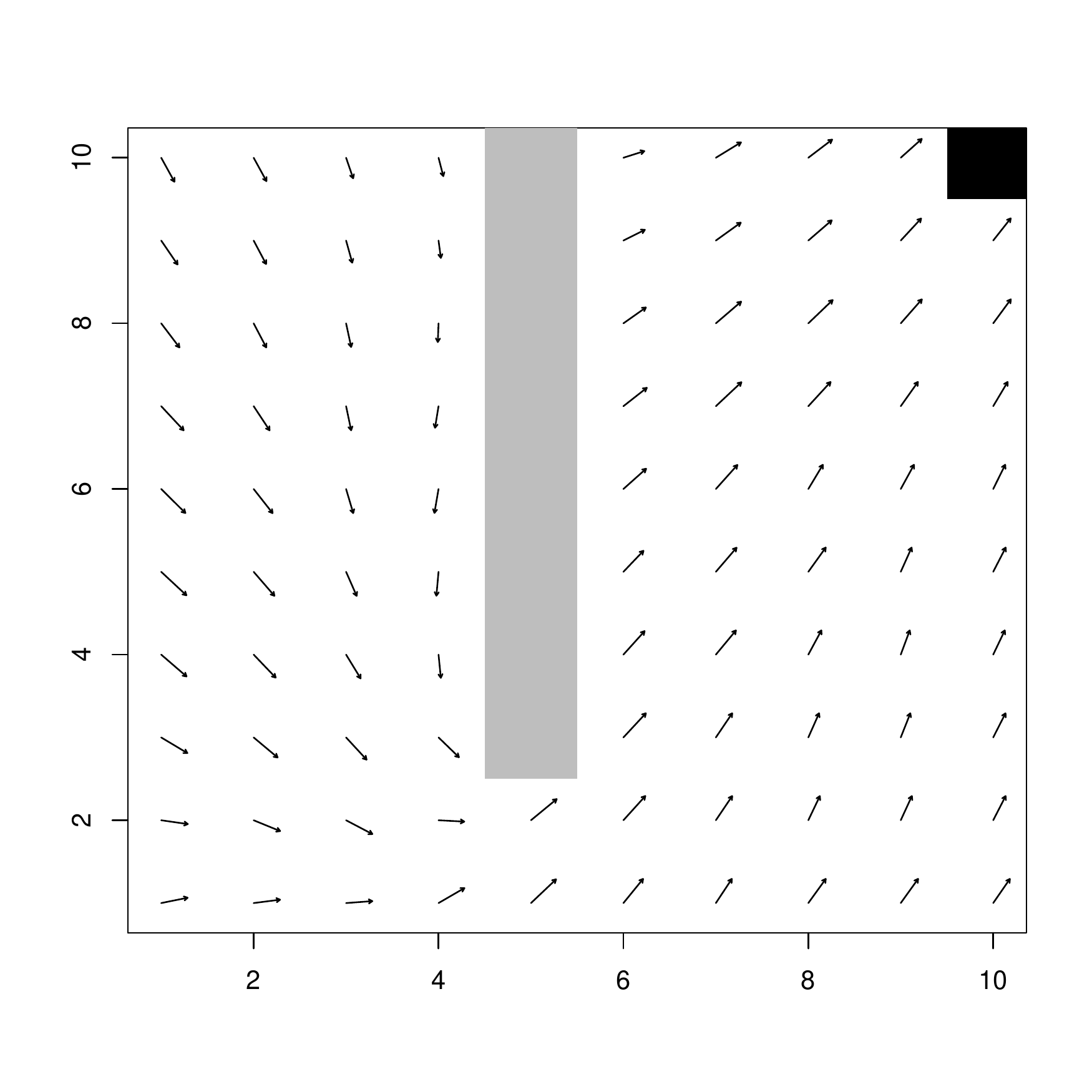}
\caption{Behavior evolved in the One-Wall Maze environment.}
\label{be2}
\end{figure}

Figure~\ref{be2} shows the behavior of the agent in the One-Wall Maze environment.
This figure was obtained in the same manner of Figure~\ref{be1}.

One-Wall Maze is a difficult problem. 
Even so, the behavior observed is accurate and stable.
This is justifiable by the SOM population capability of first self-organizing to the environment's input distribution.
Projecting and dividing the input space respecting its topology.
And secondly, the ability of developing different behavior at different parts of the space (niches) with their own subpopulation.

\begin{figure}
\centering
\includegraphics[height=2.5in]{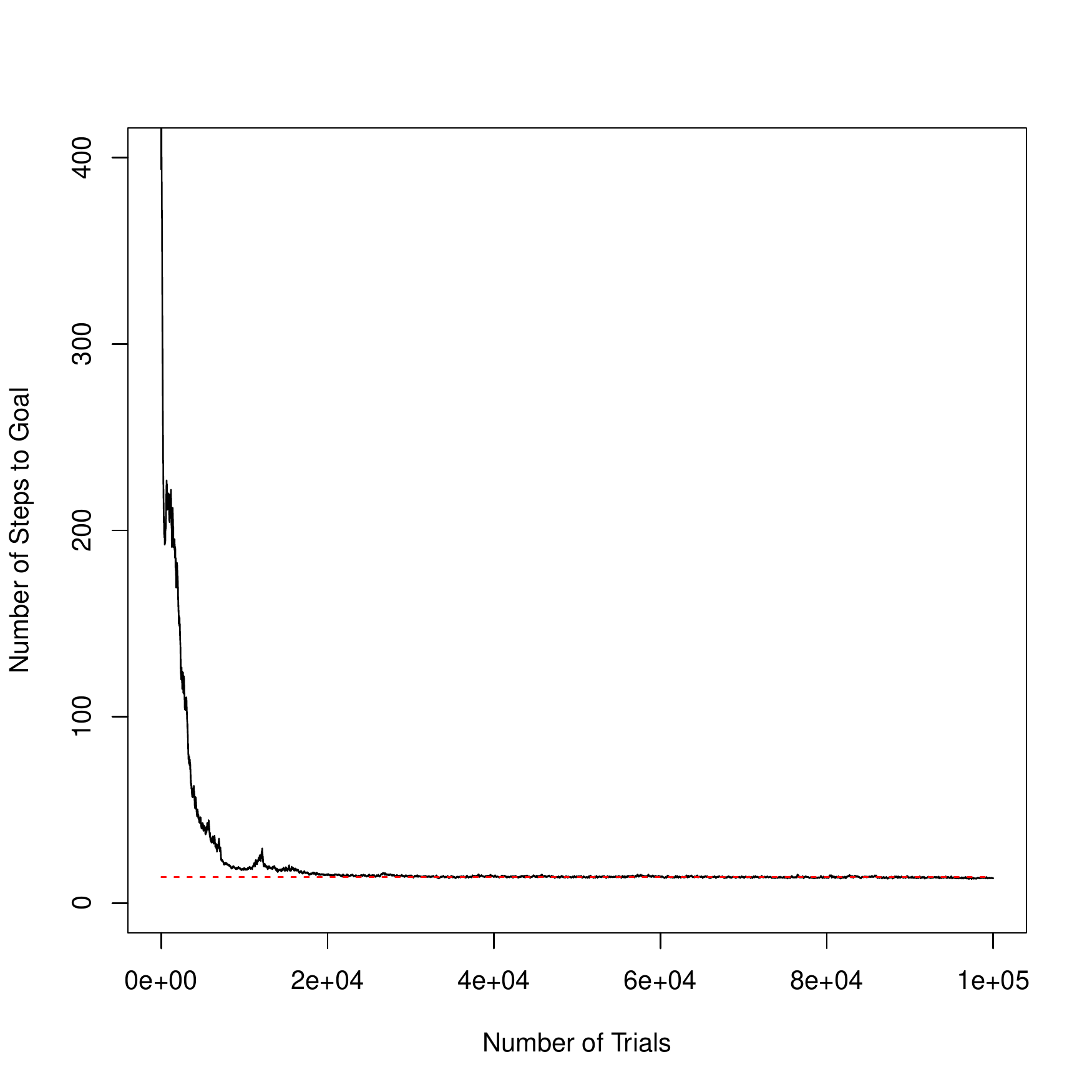}
\caption{Performance over the One-Wall Maze environment (continuous line). Dashed-line is drawn for guiding purposes parallel to the x-coordinate at $14$ of distance.}
\label{pe2}
\end{figure}
\begin{figure}
\centering
\includegraphics[height=2.5in]{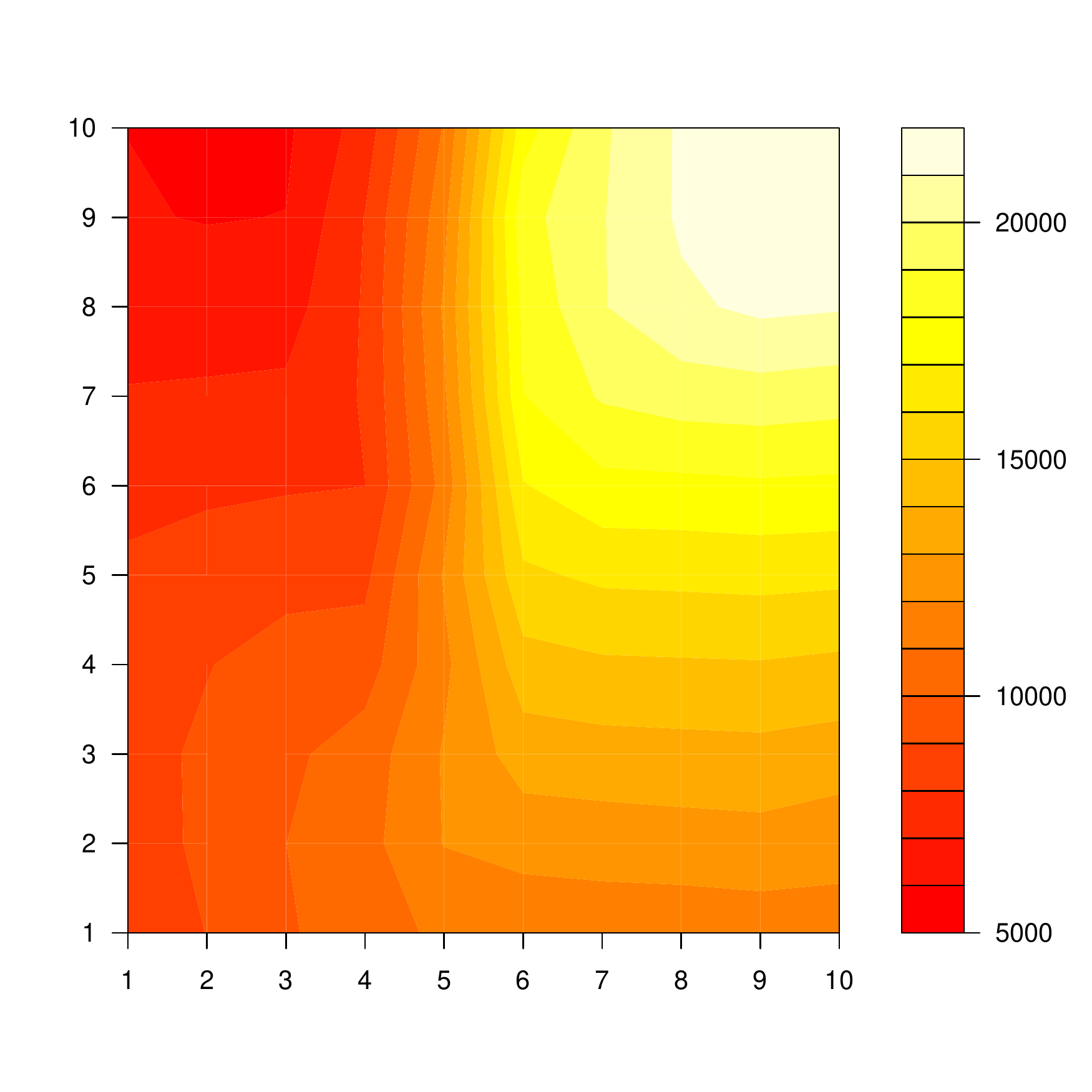}
\caption{Average distribution of fitness in the One-Wall Maze environment.}
\label{fit2}
\end{figure}

Note that the SOM used was initialized to values between zero and one as weight vectors.
Therefore, it was not in any way biased to the environment's inputs range (environment values range from zero to ten).
Additionally, Figures~\ref{pe2} and~\ref{fit2} show respectively the performance and fitness distribution.

\subsection{Population Analysis}

But what is the cost of running a SOC?
Solving a problem is often not enough, it is required that a method uses reasonably its resources.
The aim of this section is to evaluate briefly the population requirements of SOC.
All tests are run over the One-Wall Maze.

\begin{figure}
\centering
\includegraphics[height=3in]{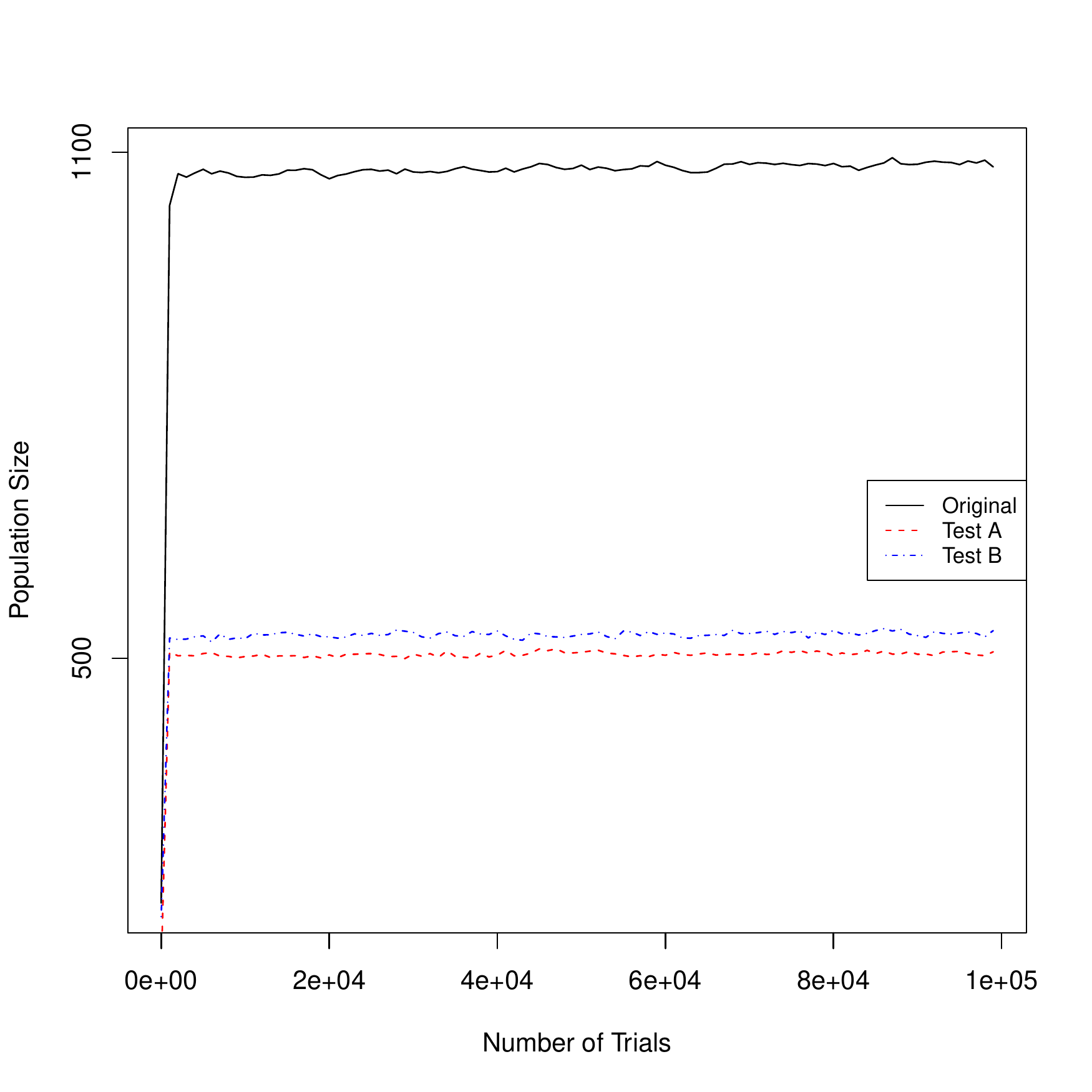}
\caption{Population dynamics for the original and two additional tests with smaller maximum populations.}
\label{pop}
\end{figure}

Foremost, SOC's maximum possible population size can be computed by the following equation:
\begin{equation}
Maximum Population Size = cells*(\beta + \nu),
\end{equation}
where $cells$ is the number of cells present in the SOM map.
Taking the parameters found in Table~\ref{para} it is possible to compute a maximum population size of $1500$ for the previous experiments.
However, as shown in Figure~\ref{pop} this value is never reached.
The population grows rapidly to around $1100$ and remains stable below this value. 

Moreover, Figure~\ref{pop} shows the population growth of two other variations of the same algorithm.
Test A has $\beta=2$ and $\nu=5$ resulting in a maximum population size of $700$, while Test B has a $7\times7$ SOM map and therefore having $735$ of maximum population size.
Note that they reach around $500$ individuals and after that the population size remains approximately constant.

\begin{figure}
\centering
\includegraphics[height=2in]{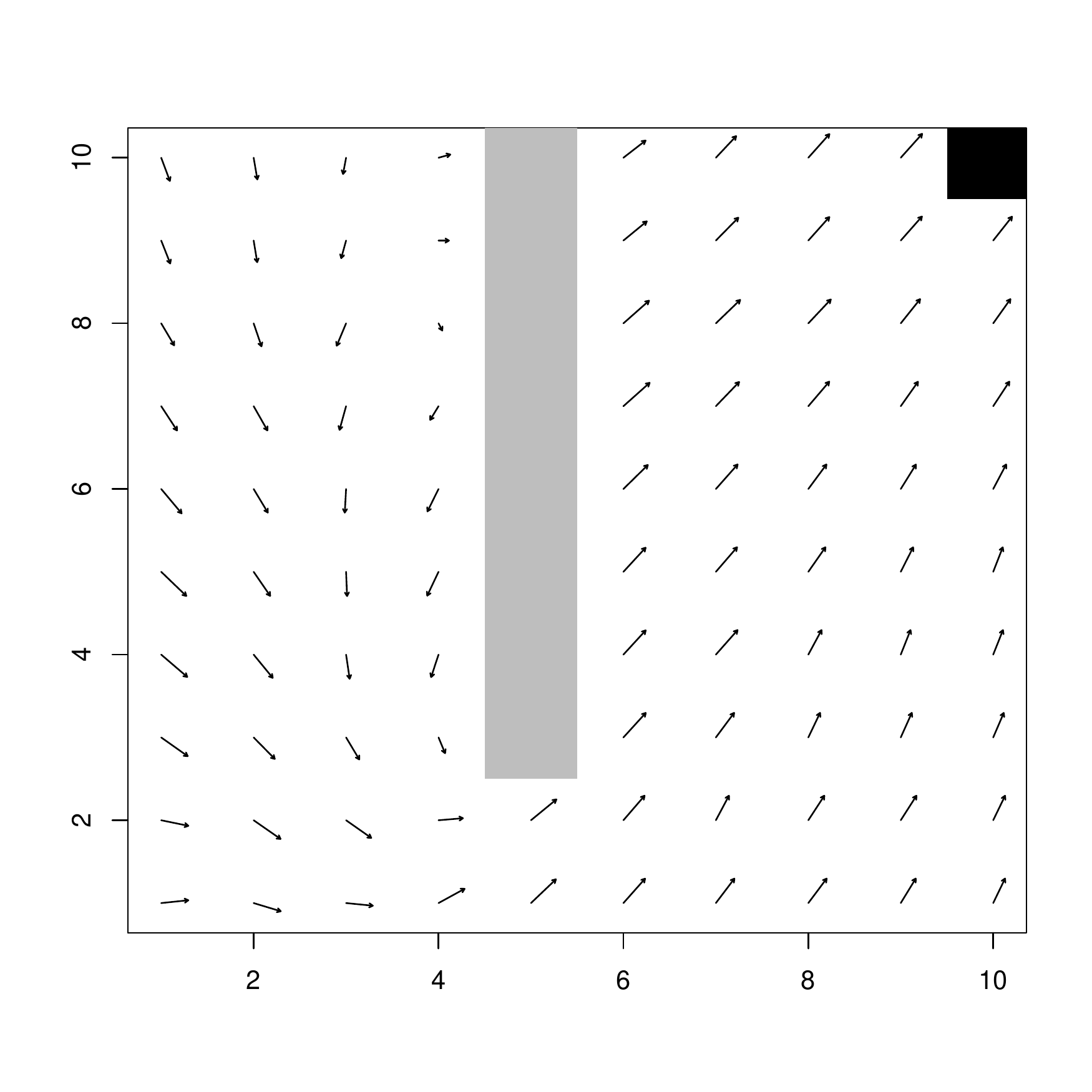}
\caption{Behavior for the algorithm with a smaller population, i.e., $\beta=2$ and $\nu=5$ (Test A).}
\label{beh_pop}
\end{figure}

\begin{figure}
\centering
\includegraphics[height=2in]{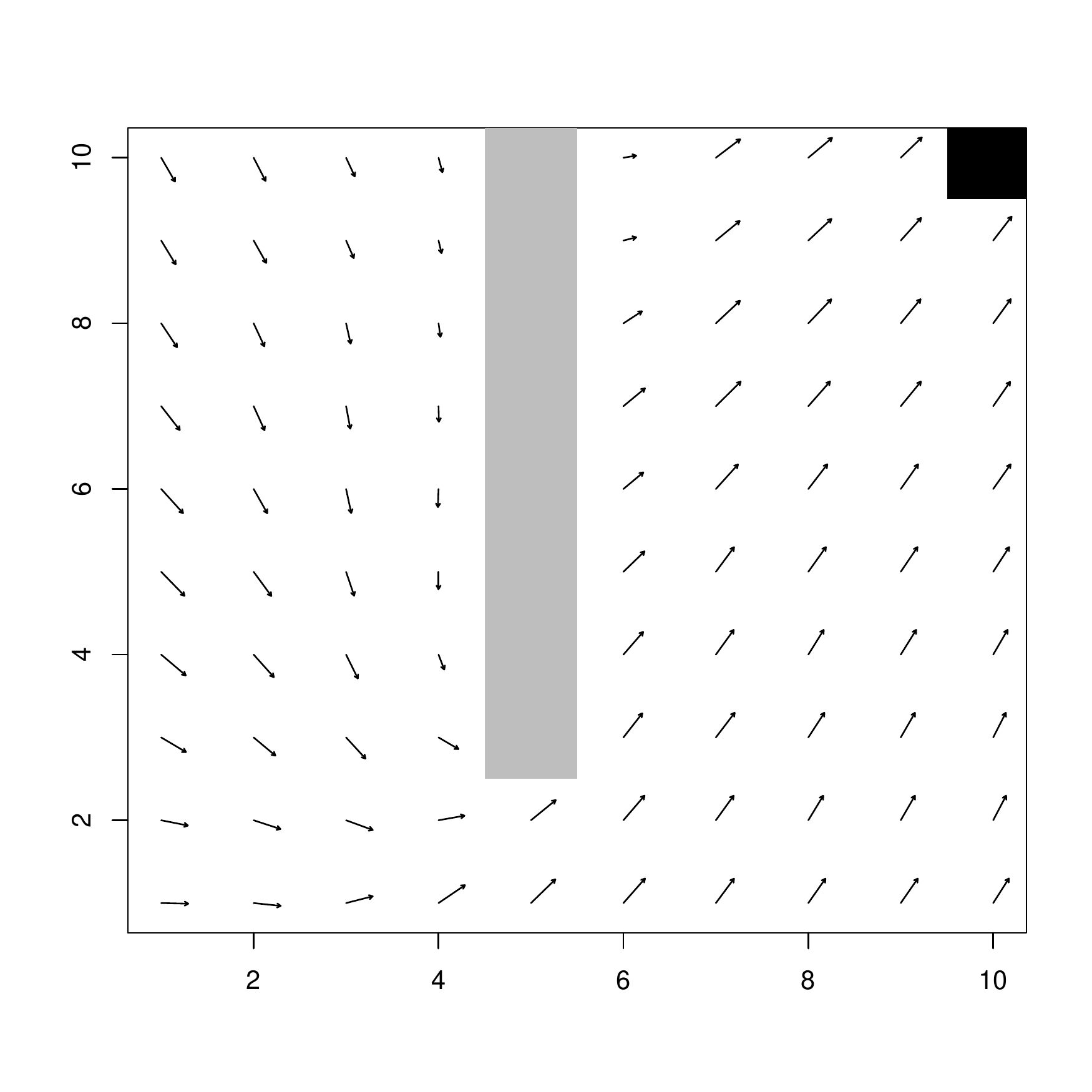}
\caption{Behavior for the algorithm with a smaller population, i.e., a $7\times7$ SOM population (Test B).}
\label{beh_7}
\end{figure}

Figures~\ref{beh_pop} and~\ref{beh_7} show respectively the behavior of Test A and Test B.
Both achieve similar actions to the original algorithm with a bigger population. 
Although, Test A have places with high direction variations which result in vectors with smaller magnitude.
This happens because smaller population sizes enable frequently accessed cells to evolve faster which makes even more infrequent the access of spots near the wall. 
Consequently, the SOM may not have any cells mapping those infrequent spots near the wall.

\begin{figure}
\centering
\includegraphics[height=3in]{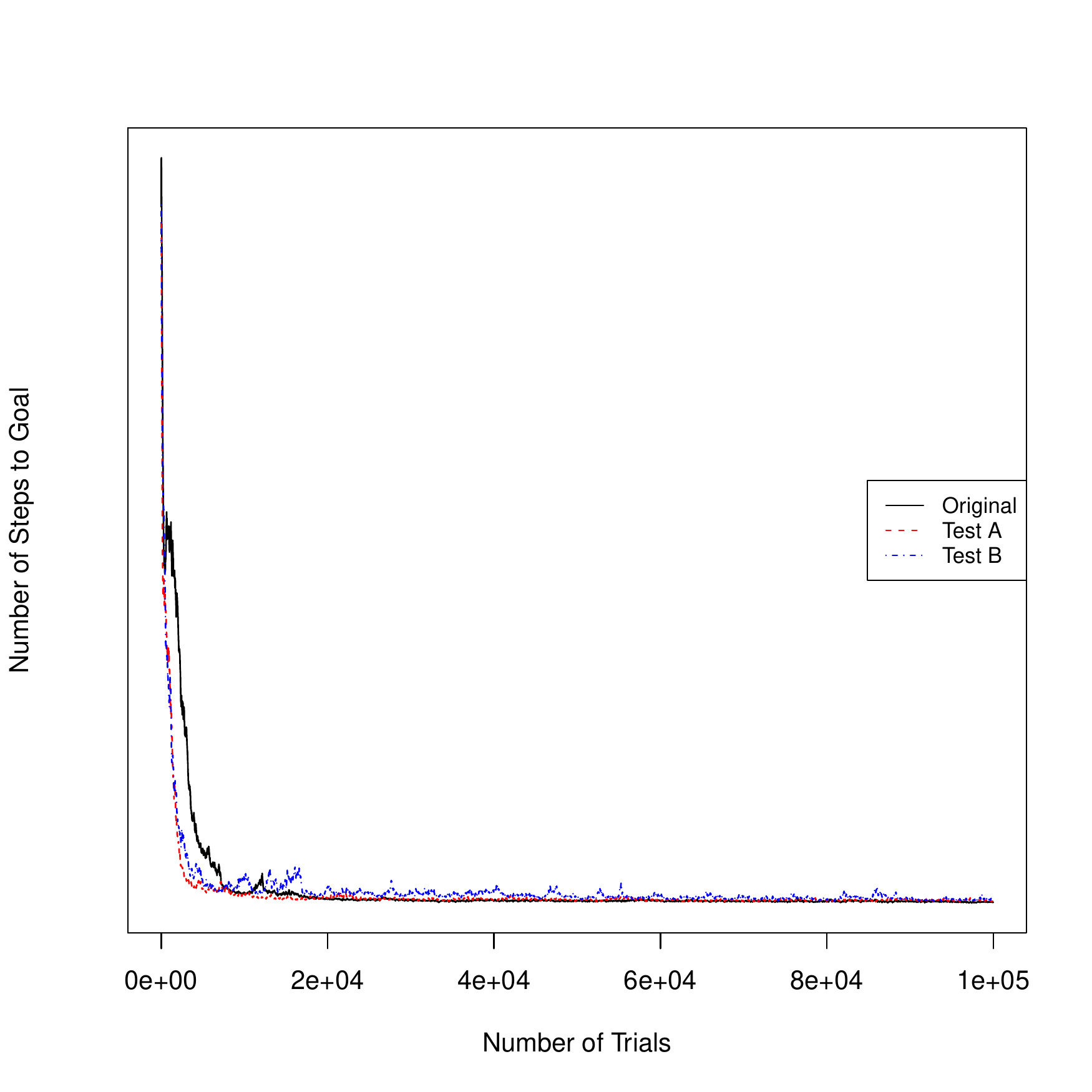}
\caption{Performance comparison between the original parameters and the smaller population versions (Tests A and B).}
\label{perfcomp}
\end{figure}

Tests A and B performance is also similar (see Figure~\ref{perfcomp}). 
Actually, Tests A and B achieves the optimum faster than the original, because their populations are smaller, however, Test B presents small deviations from the optimum.

Thus, the cost in terms of population size is small. 
Besides, as shown above, smaller populations have similar results.
Recall that a neural based LCS needed a maximum population size of $16000$ individuals to solve a continuous Empty Room problem \cite{howard2009towards}.
And XCS required on average $404$ macroclassifiers to solve a similar problem though with \textbf{discrete output} called Side 10 \cite{lanzi2005xcs}.
Therefore, SOC's results are promising when compared with the literature.

\section{Conclusions}

This paper proposed the algorithm named Self Organizing Classifiers which possess a different perspective on how to face the balance of generalized and specialized classifiers.

We showed that with a self-organizing structured population, the separation of niching and fitness pressures is possible.
Experiments on continuous multi-step problems showcased the capabilities of the approach.
Note that this article presented a problem which is more difficult than previous continuous multi-step problems.
This is just the first article about this type of algorithm and the difficulty of the problems solved shows a promising horizon for this line of algorithms.

Additionally, two new concepts were introduced:
\begin{itemize}
	\item Niched Fitness - a fitness relative to the niche;
	\item SOM Population - a dynamical structured population based on the SOM's dynamics.
\end{itemize}
Both concepts are fundamental to understand the SOC algorithm.

Thus, with the promising results accomplished by the novel concepts and the algorithm introduced here, they will, hopefully, aid the development of other algorithms and theories specially in the area of genetics based machine learning or evolutionary computation in general.



%
\bibliographystyle{abbrv}
\bibliography{sigproc}  
%
%
\end{document}